\documentclass[conference]{IEEEtran}
\IEEEoverridecommandlockouts
\usepackage{cite}
\usepackage{amsmath,amssymb,amsfonts}
\usepackage{algorithmic}
\usepackage{graphicx}
\usepackage{eso-pic} 
\usepackage{multirow}
\usepackage{textcomp}
\usepackage[table]{xcolor}
\def\BibTeX{{\rm B\kern-.05em{\sc i\kern-.025em b}\kern-.08em
    T\kern-.1667em\lower.7ex\hbox{E}\kern-.125emX}}
\definecolor{topone}{RGB}{255,153,153}
\definecolor{toptwo}{RGB}{252,220,185}
\definecolor{topthree}{RGB}{255,247,191}
\begin{document}

\title{HICA: Hierarchical Temporal Clustering with Scene-Dynamic Compatibility for Consistent Video Anomaly Detection\\
\thanks{\IEEEauthorrefmark{1}Corresponding author.}}

\author{
\IEEEauthorblockA{
\begin{tabular}{cc}
\begin{minipage}{0.42\textwidth}
\centering
Jiye Won\\
\textit{School of Computer Science \& Engineering}\\
\textit{Kyungpook National University}\\
Daegu, South Korea\\
snowball88@knu.ac.kr
\end{minipage}
&
\begin{minipage}{0.42\textwidth}
\centering
Jinhyeong Park\\
\textit{School of Computer Science \& Engineering}\\
\textit{Kyungpook National University}\\
Daegu, South Korea\\
hini2245@knu.ac.kr
\end{minipage}
\\[4em]
\multicolumn{2}{c}{
\begin{minipage}{0.42\textwidth}
\centering
Soon Ki Jung\IEEEauthorrefmark{1}\\
\textit{School of Computer Science \& Engineering}\\
\textit{Kyungpook National University}\\
Daegu, South Korea\\
skjung@knu.ac.kr
\end{minipage}
}
\end{tabular}
}
}

\AddToShipoutPictureFG*{%
  \AtPageLowerLeft{%
    \put(48,28){%
      \parbox[b]{516pt}{%
        \centering\fontsize{6.2}{7.3}\selectfont
        \textcopyright{} 2026 IEEE. Personal use of this material is permitted.
        Permission from IEEE must be obtained for all other uses, in any current or future
        media, including reprinting/republishing this material for advertising or
        promotional purposes, creating new collective works, for resale or redistribution
        to servers or lists, or reuse of any copyrighted component of this work in other works.
      }%
    }%
  }%
}

\maketitle

\begin{abstract}
Weakly supervised video anomaly detection requires frame-level precision without frame-level annotations, yet existing methods struggle to produce consistent predictions across anomaly classes. This is because current approaches process video segments through attention mechanisms that operate either locally within fixed frame grids or globally across all segments without content-specific grouping. As a result, anomaly signals become fragmented and diluted by surrounding normal frames, leading to suboptimal frame-level predictions. To address this, we propose a Hierarchical Content-Adaptive Temporal Module and a Static-Dynamic Compatibility Module for content-adaptive temporal grouping and scene-action compatibility modeling. Experiments demonstrate that our method consistently improves binary anomaly detection over the baseline, while also achieving lower inter-class variance, confirming that the proposed modules provide complementary benefits for more robust and consistent video anomaly detection.
\end{abstract}


\section{Introduction}
Video anomaly detection (VAD) requires frame-wise precise and consistent labelling of anomalies. However, current approaches struggle to accurately locate anomalies, leading to inconsistent performances across anomaly classes. Imagine a public site where surveillance is operating. If a VAD model's estimation is inconsistent on specific anomalies, for example, assault or theft, malicious individuals may exploit these weaknesses. An ideal VAD system should detect various anomaly types consistently.\\
To address this limitation and enhance the robustness of VAD, we revisit how temporal information is modeled. Recent approaches process video segments through attention~\cite{vaswani2017attention} mechanisms that operate either by matching frames at fixed positional indices across segments, or by mixing all segments globally without content-specific grouping. While these designs enable information exchange, they fail to organize segments into semantically coherent groups: positionally matched attention treats each segment independently rather than as a whole, while indiscriminate global mixing blends anomalous and normal segments into a single undifferentiated representation. As a result, anomaly signals become fragmented and diluted by surrounding normal frames, leading to suboptimal frame-level predictions.\\
To the end, we propose a content-adaptive temporal modeling approach that first encodes video features and organizes segments into semantically coherent groups, allowing temporally related patterns to be jointly represented. The aggregated context is then propagated back to individual frames to improve frame-level anomaly estimation. In addition, we introduce a scene-action compatibility mechanism that models the relationship between global scene context and local temporal dynamics, producing a score that reflects the likelihood of an action to happen in a scene. The final anomaly prediction is obtained by combining the frame-wise anomaly score with this compatibility measure, leading to more consistent and context-aware detection.\\
Specifically, our contributions can be summarized as follows: 
\begin{itemize}
    \item We introduce a hierarchical content-adaptive temporal module that organizes video segments into semantically coherent groups and models their global dependencies for consistent and precise frame-level predictions.
    \item We propose a static-dynamic compatibility module that explicitly captures the relationship between global scene context and local temporal dynamics to enhance anomaly detection consistency.
    \item We present a unified framework that integrates the two modules for more robust and context-aware video anomaly detection.
\end{itemize}

\section{Related Work}

\textbf{Weakly Supervised VAD and Temporal Modeling.}
Multiple instance learning (MIL) has been studied for weakly supervised VAD by treating videos as bags of segments and training with video-level labels alone~\cite{ucfcrime}. Following works improved robustness through temporal convolutions~\cite{zhang2019tcnibloss}, feature magnitude learning~\cite{tian2021rtfm}, and graph-based label denoising~\cite{zhong2019gcncleaner}. The integration of CLIP~\cite{radford2021learning} further advanced the field: temporal self-attention over score-weighted snippets was proposed in CLIP-TSA~\cite{joo2023cliptsa}, and normality-guided pseudo-label generation was introduced in~\cite{yang2024normality}. Following this line of work, we also adopt the MIL framework under video-level weak supervision, and build on AnomalyCLIP~\cite{zanella2024delving}, which re-centers the CLIP feature space around a normality prototype and employs an Axial Transformer~\cite{ho2019axial} for temporal modeling.

\section{Method}

\begin{figure*}[!t]
\centering
\includegraphics[width=0.85\textwidth]{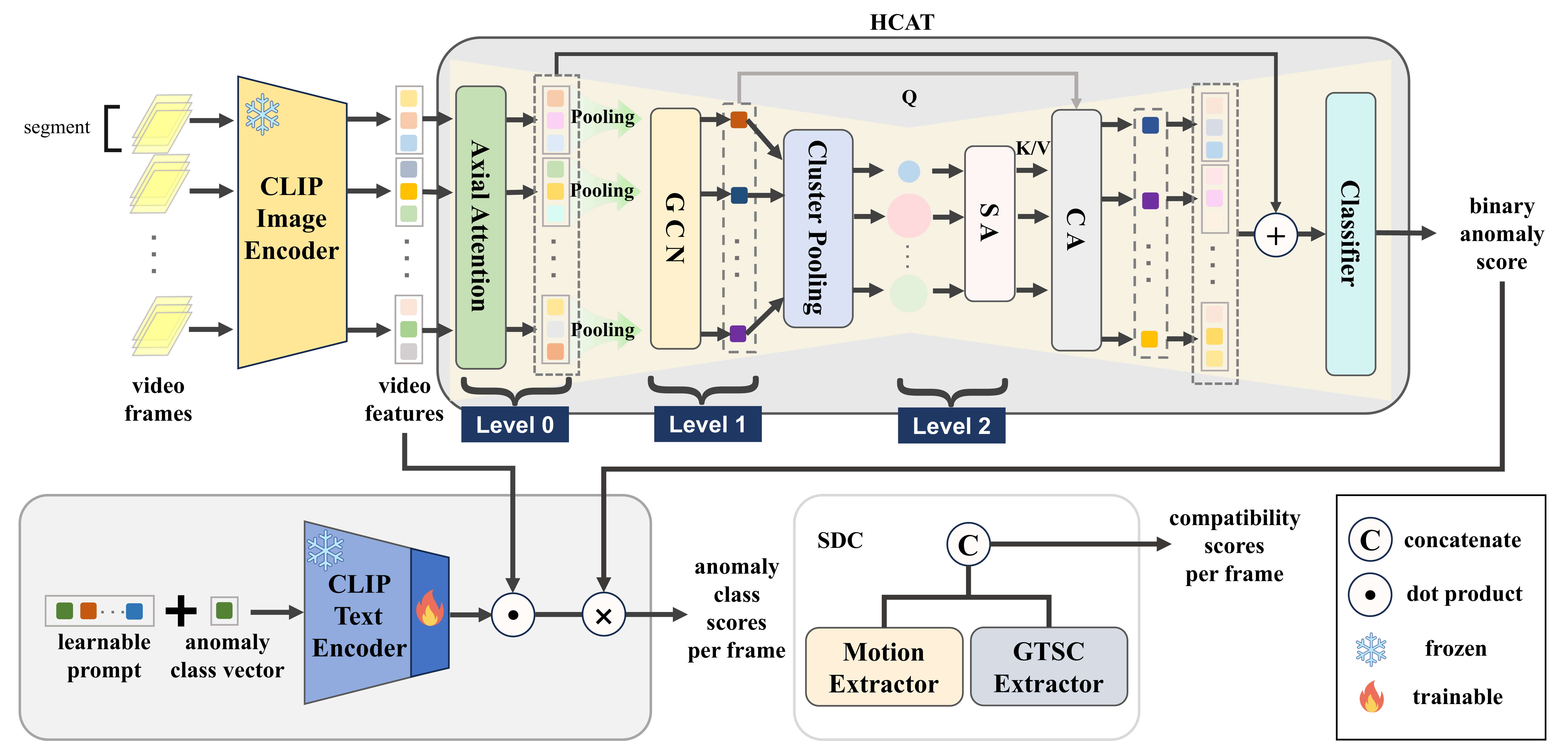}
\caption{Pipeline of the proposed method. (1) Video frames are encoded and divided into segments. (2) HCAT hierarchically clusters segments and propagates their contextual information back to frames to estimate anomaly scores. (3) SDC extracts global scene features and local motion cues to compute a compatibility score per frame. (4) The final anomaly score is obtained by combining temporal anomaly and scene-dynamic compatibility, emphasizing events that are both temporally irregular and contextually implausible.}
\label{fig:method}
\end{figure*}

In this work, we address weakly supervised video anomaly detection under the MIL framework. Each training video carries only a single binary label $y \in \{0,1\}$ indicating whether any anomaly occurs; no frame-level annotation is provided. At inference, the model outputs a per-frame anomaly score $\mathbf{s}_{t} \in [0,1]$.

\subsection{Overall Framework}

\textbf{Input Preparation.} A video with an arbitrary length is uniformly divided into $N$ non-overlapping segments of $L$ consecutive frames. A frozen CLIP image encoder extracts frame-level features $\mathbf{F} \in \mathbb{R}^{T \times D}$, where $T = N \times L$. A dataset-specific normal centroid $\boldsymbol{\mu}$, defined as the mean feature of all training normal videos, is subtracted to obtain re-centered features $\tilde{\mathbf{F}} = \mathbf{F} - \boldsymbol{\mu}$, amplifying deviations from typical normal appearance~\cite{zanella2024delving}. 

\textbf{Branch 1: Class-Logit Branch.} This branch answers what type of anomaly is present by aligning visual features with text semantics. Learnable class-conditional text prompts, optimized via CoOp~\cite{zhou2022learning}, are encoded by a CLIP text encoder into $C$ class direction vectors.
The cosine similarity between each frame feature and these directions yields class logits $\mathbf{L} \in \mathbb{R}^{T \times C}$, which serve as proxy anomaly scores. The MIL selector uses these logits to identify the $k$ most and $k$ least anomalous segments, which anchor all training loss terms.

\textbf{Branch 2: Temporal-Score Branch.} This branch answers how anomalous each frame is through dedicated temporal modeling. A shared linear projection maps $\tilde{\mathbf{F}}$ to a compact representation $\mathbf{P} \in \mathbb{R}^{T \times d}$, which is processed by our two proposed modules. HCAT (Section~\ref{sec:hcat}) builds a hierarchical temporal representation through content-adaptive grouping and top-down context propagation, producing base anomaly scores $\mathbf{s}_{\mathrm{hcat}} \in [0,1]^T$. SDC (Section~\ref{sec:sdc}) independently measures the frame-level compatibility between the stable background scene and the dynamic foreground motion, producing a compatibility score $\mathbf{c} \in [0,1]^T$. The two are composed via score modulation:
\begin{equation}
\mathbf{s}_{\mathrm{final}}=\mathbf{s}_{\mathrm{hcat}}\odot\bigl(1+\lambda(1-\mathbf{c})\bigr)
\end{equation}
Frames that are simultaneously temporally anomalous, corresponding to high $\mathbf{s}_{\mathrm{hcat}}$, and scene-incompatible, corresponding to low $\mathbf{c}$, receive amplified final scores.

\textbf{Inference Output.} At inference, the two branches jointly determine the final predictions. $\mathbf{s}_{\mathrm{final}}$ from Branch 2 serves as the binary anomaly score. The multi-class anomaly score is obtained by weighting the class-conditional probabilities from Branch 1 by the binary score from Branch 2:
\begin{equation}
\mathbf{P}_{\mathrm{class}}=\mathrm{softmax}(\mathbf{L})\odot\mathbf{s}_{\mathrm{final}}.
\end{equation}

\subsection{Hierarchical Content-Adaptive Temporal Model (HCAT)}\label{sec:hcat}

Effective temporal modeling for VAD requires more than information exchange across segments. This requires organizing segments into semantically coherent groups so that anomalous patterns can be jointly represented without being diluted by surrounding normal frames. Content-adaptive grouping provides a fundamentally different inductive bias from uniform attention: rather than producing a soft mixture over all segments, it compresses semantically similar segments into compact and isolated cluster representations, enabling the model to capture coherent anomalous events regardless of their temporal position or duration.

HCAT realizes this through a three-level hierarchy: it first compresses frames into segment-level holistic vectors, then groups segments into content-adaptive clusters via cross-attention, and finally propagates the resulting cluster-level context back to each individual frame:
\begin{equation}
\underbrace{\mathrm{Frame}}_{\mathrm{Level~0}}\xrightarrow{\mathrm{pool}}\underbrace{\mathrm{Segment}}_{\mathrm{Level~1}}\xrightarrow{\mathrm{cluster}}\underbrace{\mathrm{Cluster}}_{\mathrm{Level~2}}\xrightarrow{\mathrm{broadcast}}\underbrace{\mathrm{Frame}}_{\substack{\mathrm{Level~0}\\(\mathrm{enhanced})}}.
\end{equation}

\subsubsection{Frame-Level Encoding via Axial Attention (Level 0)}
Features $\tilde{\mathbf{F}}$ are linearly projected to embeddings $\mathbf{X} \in \mathbb{R}^{T \times d}$. An Axial Transformer~\cite{ho2019axial} processes $\mathbf{X}$ along two decomposed axes over the $N \times L$ segment-frame grid.

\begin{itemize}
\item \textbf{Row Attention (along $L$)}: captures dependencies between frames within the same segment, encoding intra-segment temporal context.
\item \textbf{Column Attention (along $N$)}: captures dependencies between the same temporal positions across different segments, encoding inter-segment positional context.
\end{itemize}
The output $\mathbf{X}^{(0)} \in \mathbb{R}^{N \times L \times d}$ provides frame-level representations with awareness of both local and cross-segment structure. $\mathbf{X}$ is also retained as input to the SDC module (Section~\ref{sec:sdc}).

\subsubsection{Learned-Weighted Segment Pooling (Level 0$\to$1)}
To obtain a holistic representation of each segment, the $L$ frame representations from $\mathbf{X}^{(0)}$ are compressed into a single vector via learned weighted pooling. Specifically, a linear layer computes a scalar importance score for each frame; the scores are softmax-normalized across the $L$ frames within the segment, and the resulting weighted sum is normalized by LayerNorm, yielding segment features $\mathbf{H} \in \mathbb{R}^{N \times d}$. This treats each segment as a single entity rather than a collection of independent frames, enabling the holistic inter-segment reasoning performed by the subsequent GCN.

\subsubsection{Temporal-Semantic GCN (Level 1)}
Before clustering, we enrich segment representations by propagating information across temporally and semantically related segments. Motivated by prior work on temporal graph modeling in VadCLIP~\cite{wu2024vadclip}, we design a temporal-semantic GCN whose per-video adjacency matrix combines temporal proximity and semantic similarity in a multiplicative structure:
\begin{equation}
\mathbf{A}_{ij} =
\underbrace{\exp\left(-\frac{(i-j)^2}{2\sigma^2}\right)}_{\text{temporal proximity}}
\cdot
\left(1 +
\underbrace{\mathrm{ReLU}\left(\frac{\mathbf{h}_i^\top \mathbf{h}_j}{\lVert \mathbf{h}_i \rVert \lVert \mathbf{h}_j \rVert}\right)}_{\text{semantic similarity}}
\right)
\end{equation}
where $\sigma$ is a temporal bandwidth hyperparameter and $\mathbf{h}_i, \mathbf{h}_j \in \mathbb{R}^{d}$ denote the $i$-th and $j$-th rows of $\mathbf{H}$. The temporal term keeps adjacent segments connected, while the semantic term dynamically bridges distant segments that share similar content. The multiplicative structure prevents spurious long-range connections between temporally isolated segments, while still enabling semantic amplification at moderate distances; the bandwidth $\sigma$ is tuned per dataset to control the effective temporal reach of each edge. After row-normalizing $\mathbf{A}$ into $\hat{\mathbf{A}}$, a GCN with learnable weight $\mathbf{W} \in \mathbb{R}^{d \times d}$ and residual connection yields:
\begin{equation}
\mathbf{H}^{(1)} = \mathrm{LN}(\mathbf{H} + \hat{\mathbf{A}}\mathbf{H}\mathbf{W}) \in \mathbb{R}^{N \times d}
\end{equation}

\subsubsection{Content-Adaptive Cluster Pooling (Level 1$\to$2)}
The $N$ segment representations are compressed into $K$ cluster representations via Perceiver-style cross-attention~\cite{jaegle2021perceiver}, where $K$ learnable query vectors attend over $\mathbf{H}^{(1)}$ to produce cluster features $\mathbf{C} \in \mathbb{R}^{K \times d}$. Unlike fixed temporal windows or positional pooling, this content-adaptive design has two key properties. First, grouping is driven by feature similarity rather than temporal position: each query attends to segments that share a similar semantic pattern regardless of where they appear in the video. Second, the soft attention weights adjust per video: the same query may cover many segments in a video with prolonged anomalous activity, and fewer in one with a brief localized event. Together, these properties allow the model to form a compact global representation that reflects the actual semantic structure of each video, rather than its temporal layout.

\subsubsection{Cluster Self-Attention (Level 2)}
The $K$ cluster representations undergo self-attention with a residual connection and FFN, following the standard Pre-LN block.
This enables direct mutual reasoning across the $K$ learned groupings. Since $K \ll N$, the attention operates over a compact set of representations, providing efficient global reasoning without the noise of full $N \times N$ segment-level attention.

\subsubsection{Cross-Level Attention and Hierarchical Broadcast}
The cluster-level context is propagated back to frame level in two stages. First, each segment attends to the $K$ cluster representations via cross-attention, absorbing the global semantic structure captured at the cluster level. This yields enhanced segment features $\mathbf{H}^{(2)} \in \mathbb{R}^{N \times d}$. Second, $\mathbf{H}^{(2)}$ is broadcast uniformly to all $L$ frames within each segment via residual addition.
\begin{equation}
\mathbf{X}^{(2)}_{n,l} = \mathbf{X}^{(0)}_{n,l} + \mathrm{LN}(\mathbf{H}^{(2)}_n), \quad \forall\, l \in \{1,\ldots,L\}.
\end{equation}
The final HCAT output $\mathbf{X}^{(2)} \in \mathbb{R}^{N \times L \times d}$ integrates three levels of context: frame-level structure from axial attention ($\mathbf{X}^{(0)}$), relational enrichment from the semantic GCN ($\mathbf{H}^{(1)}$), and global cluster context from the hierarchical encoding ($\mathbf{C}^{(2)}$). These features are passed through a classification head to produce per-frame base anomaly scores $\mathbf{s}_{\mathrm{hcat}} \in [0,1]^T$.

\subsection{Static-Dynamic Compatibility Module (SDC)}\label{sec:sdc}
The SDC module measures the compatibility between the stable background scene context and the local dynamic motion at each frame $t \in \{1, \ldots, T\}$. When the two are inconsistent, the base anomaly score $\mathbf{s}_{\mathrm{hcat}}$ from HCAT is amplified; when motion is consistent with the scene, the score is left unchanged.

\subsubsection{Global Time-Stable Context (GTSC) Extractor}
We estimate the scene background as the temporal mean of the axial attention~\cite{ho2019axial} output $\mathbf{X}^{(0)} \in \mathbb{R}^{T \times d}$ as:
\begin{equation}
\mathbf{G} = \frac{1}{T}\sum_{t'=1}^{T} \mathbf{x}^{(0)}_{t'}, \quad \forall\ t.
\end{equation}
Therefore, $\mathbf{G} \in \mathbb{R}^{T \times d}$ is a time-independent background signature.

\subsubsection{Dual-Path Motion Extractor}
Motion is characterized along two complementary paths and then fused. The first path captures low-level appearance change by computing frame-to-frame feature differences from $\mathbf{X}$---the linear projection retained prior to axial attention---preserving the raw signal before temporal smoothing. These differences are encoded by a two-layer 1D convolution to produce $\mathbf{m}_{\text{low}} \in \mathbb{R}^{T \times d/2}$. The second path extracts semantically contextualized motion by applying a linear projection over the axial attention~\cite{ho2019axial} output $\mathbf{X}^{(0)}$, yielding $\mathbf{m}_{\text{sem}} \in \mathbb{R}^{T \times d/2}$. The two are concatenated and fused via a linear layer to produce $\mathbf{M}_{t} \in \mathbb{R}^{T \times d}$. Path 1 captures abrupt transitions that Path 2 may miss; Path 2 provides semantic context that raw differences cannot convey.

\subsubsection{Compatibility Scoring and Score Modulation}
$\mathbf{G}_t$ and $\mathbf{M}_t$ are jointly projected and passed through a lightweight MLP to produce a per-frame compatibility score $c_t \in [0, 1]$:
\begin{equation}
 c_t = \sigma \left( \mathrm{MLP} \left( [ \mathbf{W}_{\text{ctx}} \mathbf{G}_t ; \mathbf{W}_{\text{mot}} \mathbf{M}_t ] \right) \right)
\end{equation}
where $c_t \approx 1$ indicates motion compatible with the background and $c_t \approx 0$ signals a mismatch. The final anomaly score is obtained by multiplying $s_t$ by $\left(1 + \lambda(1 - c_t)\right)$, which amplifies the signal when compatibility is low.

\subsection{Training Objective}

Let $\mathcal{V}^+ = \{V^+_1, \ldots, V^+_k\}$ denote the $k$-most anomalous segments and $\mathcal{V}^- = \{V^-_1, \ldots, V^-_k\}$ the $k$-least anomalous segments selected from an abnormal video of class $c$, and $\mathcal{S}^+ = \{S^+_1, \ldots, S^+_k\}$ the $k$-most anomalous segments selected from a normal video. $L_{t,c}$ denotes the class logit for frame $t$ and class $c$ from $\mathbf{L}$, and $p_{A,c}$ and $p_N$ denote the probability of a frame being anomalous of class $c$ and normal, respectively, derived from $\mathbf{P}_{\text{class}}$.

$\mathcal{L}_{\text{dir}}$ provides directional alignment in the CLIP feature space: it maximizes the class-$c$ logit magnitude for the $k$-most anomalous segments, while minimizing the maximum class logit across all classes for normal video segments:
\begin{equation}
\mathcal{L}_{\text{dir}} = -\frac{1}{kF}\sum_{i=1}^{k}\sum_{j=1}^{F} L_{V^+_{i,j}, c} + \frac{1}{NF}\sum_{i=1}^{N}\sum_{j=1}^{F} \max_{c} L_{S_{i,j}, c}.
\end{equation}

$\mathcal{L}_{\text{topk}}$ is the sum of two cross-entropy terms: it pushes the top-$k$ segments of abnormal videos toward their ground-truth class $c$, and the top-$k$ segments of normal videos toward the normal class:

\begin{equation}
    \mathcal{L}_{\text{topk}} = -\frac{1}{kF}\sum_{i=1}^{k}\sum_{j=1}^{F}(\log p_{A,c}(V^+_{i,j})-\log p_N(S^+_{i,j})).
\end{equation}

$\mathcal{L}_{\text{bottomk}}$ pushes the $k$-least anomalous segments of abnormal videos toward the normal class, preventing the model from assigning uniformly high scores:
\begin{equation}
\mathcal{L}_{\text{bottomk}} = -\frac{1}{kF}\sum_{i=1}^{k}\sum_{j=1}^{F}\log p_N(V^-_{i,j}).
\end{equation}

$\mathcal{L}_{\text{smooth}}$ penalizes abrupt score changes between adjacent frames, enforcing temporal continuity:
\begin{equation}
\mathcal{L}_{\text{smooth}} = \sum_{t=2}^{T}(s_t - s_{t-1})^2.
\end{equation}

$\mathcal{L}_{\text{sparse}}$ applies an $\ell_1$ penalty on the anomaly scores of abnormal videos, reflecting the prior that anomalous frames are a minority:
\begin{equation}
\mathcal{L}_{\text{sparse}} = \frac{1}{T}\sum_{t=1}^{T} s_t.
\end{equation}

The overall training loss combines all five terms and two loss weight terms $\lambda_{\text{sm}}$ and $\lambda_{\text{sp}}$ as:
\begin{equation}
\label{eq:loss}
\mathcal{L} = \mathcal{L}_{\text{dir}} + \mathcal{L}_{\text{topk}} + \mathcal{L}_{\text{bottomk}} + \lambda_{\text{sm}}\mathcal{L}_{\text{smooth}} + \lambda_{\text{sp}}\mathcal{L}_{\text{sparse}}.
\end{equation}

\begin{table*}[!t]
\caption{Comparison of weakly supervised methods across ShanghaiTech, UCF-Crime, and XD-Violence.}
\label{tab:main-results}
\begin{center}
\scriptsize
\begin{tabular}{|l|l|c|c|cc|cc|cc|}
\hline
\multirow{2}{*}{\textbf{Method}} & \multirow{2}{*}{\textbf{Features}} & \multirow{2}{*}{\textbf{VAD}} & \multirow{2}{*}{\textbf{VAR}} & \multicolumn{2}{c|}{\textbf{UCF-Crime}} & \multicolumn{2}{c|}{\textbf{ShanghaiTech}} & \multicolumn{2}{c|}{\textbf{XD-Violence}} \\
\cline{5-10}
 & & & & \textbf{AUC (\%)} & \textbf{mAUC (\%)} & \textbf{AUC (\%)} & \textbf{mAUC (\%)} & \textbf{AP (\%)} & \textbf{mAP (\%)} \\
\hline
Wu et al.~\cite{wu2020not} & I3D-RGB & \checkmark &  & 82.44 & -- & -- & -- & 73.20 & -- \\
RTFM~\cite{tian2021rtfm} & I3D-RGB & \checkmark &  & 84.03 & 84.86 & 97.21 & 81.60 & 77.81 & 43.04 \\
Wu and Liu~\cite{wu2021learning} & I3D-RGB & \checkmark &  & 84.89 & -- & 97.48 & -- & 75.90 & -- \\
MSL~\cite{li2022self} & I3D-RGB & \checkmark &  & 85.30 & -- & 96.08 & -- & 78.28 & -- \\
MSL~\cite{li2022self} & VideoSwin-RGB & \checkmark &  & 85.62 & -- & 97.32 & -- & 78.58 & -- \\
S3R~\cite{wu2022self} & I3D-RGB & \checkmark &  & 85.99 & 86.55 & 97.48 & 87.88 & 80.26 & 36.06 \\
MGFN~\cite{chen2023mgfn} & I3D-RGB & \checkmark &  & 86.98 & -- & -- & -- & 79.19 & -- \\
MGFN~\cite{chen2023mgfn} & VideoSwin-RGB & \checkmark &  & 86.67 & -- & -- & -- & 80.11 & -- \\
ActionCLIP~\cite{wang2021actionclip} & ViT-B/16 &  & \checkmark & 82.30 & 87.72 & 96.36 & 75.63 & 61.01 & 40.24 \\
AnomalyCLIP~\cite{zanella2024delving} & ViT-B/16 & \checkmark & \checkmark & 86.36 & 90.66 & 98.07 & 96.46 & 78.51 & \textbf{49.41} \\
\hline
Ours & ViT-B/16 & \checkmark & \checkmark & \textbf{87.42} & \textbf{91.12} & \textbf{98.52} & \textbf{96.74} & \textbf{80.39} & 48.57 \\
\hline
\end{tabular}%
\end{center}
\end{table*}

\begin{table*}[!t]
\caption{Per-class AUC comparison on UCF-Crime.}
\label{tab:ucf-class-results}
\begin{center}
\scriptsize
\setlength{\tabcolsep}{3pt}
\resizebox{\textwidth}{!}{%
\begin{tabular}{|l|ccccccccccccc|c|c|}
\hline
\multirow{2}{*}{\textbf{Method}} & \multicolumn{13}{c|}{\textbf{Class}} & \multirow{2}{*}{\textbf{mAUC}} & \multirow{2}{*}{$\sigma^2$} \\
\cline{2-14}
 & \textbf{Abuse} & \textbf{Arrest} & \textbf{Arson} & \textbf{Assault} & \textbf{Burglary} & \textbf{Explosion} & \textbf{Fighting} & \textbf{RoadAcc.} & \textbf{Robbery} & \textbf{Shooting} & \textbf{Shoplifting} & \textbf{Stealing} & \textbf{Vandalism} & & \\
\hline
RTFM~\cite{tian2021rtfm} & 79.99 & 62.57 & 90.53 & 82.27 & 85.53 & 92.76 & 85.21 & 90.31 & 81.17 & 82.82 & 92.56 & 90.23 & 87.20 & 84.86 & 58.73 \\
S3R~\cite{wu2022self} & 86.38 & 68.45 & 92.19 & 93.55 & 86.91 & 93.55 & 81.69 & 85.03 & 82.07 & 85.32 & 91.64 & 94.59 & 83.82 & 86.55 & 46.39 \\
SSRL~\cite{li2022scale} & 95.33 & 79.26 & 93.27 & 91.74 & 89.06 & 92.25 & 87.36 & 80.24 & 87.75 & 84.50 & 92.31 & 94.22 & 88.17 & 88.88 & 23.93 \\
CLIP zero-shot~\cite{radford2021learning} & 57.37 & 80.65 & 93.72 & 80.83 & 74.34 & 90.31 & 83.54 & 87.46 & 70.22 & 63.99 & 71.21 & 45.49 & 66.45 & 74.28 & 175.83 \\
ActionCLIP~\cite{wang2021actionclip} & 91.88 & 90.47 & 89.21 & 86.87 & 81.31 & 94.08 & 83.23 & 94.34 & 82.82 & 70.53 & 91.60 & 94.06 & 89.89 & 87.72 & 42.52 \\
AnomalyCLIP~\cite{zanella2024delving} & 75.03 & 94.56 & 96.66 & 94.80 & 90.08 & 94.79 & 88.76 & 93.30 & 86.85 & 87.45 & 89.47 & 97.00 & 89.78 & 90.66 & 31.37 \\
\hline
Ours & 80.80 & 94.90 & 95.36 & 92.66 & 93.37 & 95.17 & 93.23 & 92.39 & 88.11 & 86.19 & 87.02 & 96.83 & 88.48 & \textbf{91.12} & \textbf{19.88} \\
\hline
\end{tabular}%
}
\end{center}
\end{table*}

\begin{table*}[!t]
\caption{Per-class AUC comparison on ShanghaiTech.}
\label{tab:shanghaitech-class-results}
\begin{center}
\scriptsize
\setlength{\tabcolsep}{3pt}
\resizebox{\textwidth}{!}{%
\begin{tabular}{|l|ccccccccccccccc|c|c|}
\hline
\multirow{2}{*}{\textbf{Method}} & \multicolumn{15}{c|}{\textbf{Class}} & \multirow{2}{*}{\textbf{mAUC}} & \multirow{2}{*}{$\sigma^2$} \\
\cline{2-16}
 & \textbf{Car} & \textbf{Chasing} & \textbf{Circuit} & \textbf{Fall} & \textbf{Fighting} & \textbf{Jumping} & \textbf{Monocycle} & \textbf{Push} & \textbf{Robbery} & \textbf{Running} & \textbf{Skateboard} & \textbf{Stoop} & \textbf{ThrowingObj.} & \textbf{Vaudeville} & \textbf{Vehicle} & & \\
\hline
RTFM~\cite{tian2021rtfm} & 99.70 & 95.41 & 99.83 & 70.19 & 97.36 & 89.14 & 37.99 & 35.28 & 67.01 & 90.59 & 96.81 & 64.11 & 97.93 & 91.75 & 90.85 & 81.60 & 458.11 \\
S3R~\cite{wu2022self} & 98.71 & 96.80 & 99.97 & 85.63 & 95.93 & 69.33 & 96.82 & 54.76 & 61.19 & 94.43 & 96.92 & 75.46 & 97.63 & 97.78 & 96.84 & 87.88 & 217.40 \\
SSRL~\cite{li2022scale} & 99.35 & 97.31 & 99.95 & 91.24 & 96.88 & 93.07 & 89.74 & 90.62 & 91.81 & 94.47 & 97.73 & 71.81 & 98.44 & 96.32 & 95.49 & 93.61 & 45.42 \\
CLIP zero-shot~\cite{radford2021learning} & 61.65 & 77.88 & 5.95 & 61.73 & 79.37 & 23.68 & 77.78 & 63.36 & 37.71 & 54.39 & 76.15 & 8.47 & 44.10 & 65.97 & 27.08 & 51.02 & 603.63 \\
ActionCLIP~\cite{wang2021actionclip} & 98.50 & 93.86 & 98.59 & 16.38 & 97.45 & 89.63 & 98.05 & 8.14 & 67.36 & 78.25 & 97.10 & 0.76 & 97.70 & 98.65 & 93.97 & 75.63 & 1202.94 \\
AnomalyCLIP~\cite{zanella2024delving} & 98.08 & 96.66 & 97.97 & 96.69 & 98.03 & 95.48 & 86.89 & 97.99 & 95.00 & 97.95 & 97.29 & 98.62 & 96.50 & 96.97 & 96.79 & 96.46 & 7.75 \\
\hline
Ours & 97.81 & 96.55 & 99.84 & 97.53 & 97.01 & 94.70 & 91.70 & 97.22 & 94.87 & 97.73 & 96.77 & 97.75 & 97.31 & 96.65 & 97.61 & \textbf{96.74} & \textbf{3.66} \\
\hline
\end{tabular}%
}
\end{center}
\end{table*}

\begin{table*}[!t]
\caption{Per-class mAP comparison on XD-Violence.}
\label{tab:xdviolence-class-results}
\begin{center}
\setlength{\tabcolsep}{3pt}
\begin{tabular}{|l|cccccc|c|c|}
\hline
\multirow{2}{*}{\textbf{Method}} & \multicolumn{6}{c|}{\textbf{Class}} & \multirow{2}{*}{\textbf{mAP}} & \multirow{2}{*}{$\sigma^2$} \\
\cline{2-7}
 & \textbf{Abuse} & \textbf{CarAccident} & \textbf{Explosion} & \textbf{Fighting} & \textbf{Riot} & \textbf{Shooting} & & \\
\hline
RTFM~\cite{tian2021rtfm} & 9.25 & 25.36 & 53.53 & 61.73 & 90.38 & 18.01 & 43.04 & 796.87 \\
S3R~\cite{wu2022self} & 2.63 & 23.82 & 45.29 & 49.88 & 90.41 & 4.34 & 36.06 & 917.26 \\
CLIP zero-shot~\cite{radford2021learning} & 0.32 & 12.21 & 22.26 & 25.25 & 66.60 & 1.26 & 21.32 & \textbf{498.84} \\
ActionCLIP~\cite{wang2021actionclip} & 2.73 & 25.15 & 55.28 & 58.09 & 87.31 & 12.87 & 40.24 & 857.40 \\
AnomalyCLIP~\cite{zanella2024delving} & 6.10 & 31.31 & 68.75 & 71.44 & 92.74 & 26.13 & \textbf{49.41} & 913.65 \\
\hline
Ours & 3.81 & 32.53 & 69.13 & 68.02 & 94.00 & 23.96 & 48.57 & 955.33 \\
\hline
\end{tabular}
\end{center}
\end{table*}

\begin{figure}[!t]
\centering
\includegraphics[width=\columnwidth]{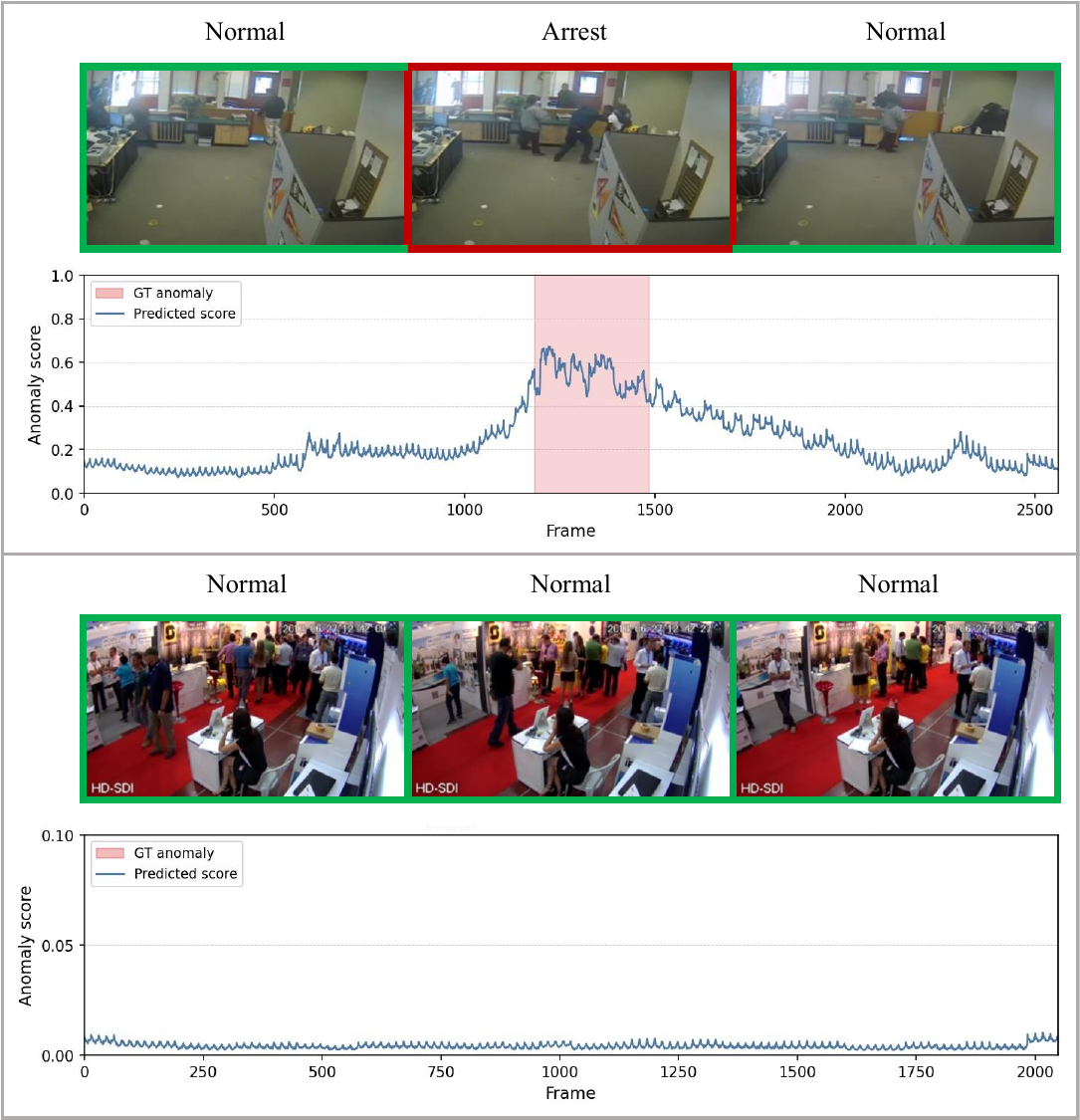}
\caption{Qualitative results on UCF-Crime. Frame-level anomaly scores predicted by our method are shown for an anomalous video (top) and a normal video (bottom). Red shading indicates ground-truth anomalous regions. Our method assigns consistently elevated scores within the annotated anomalous interval while maintaining low scores elsewhere, including across the entire normal video.}
\label{fig:qualitative}
\end{figure}

\section{Experiments}

\subsection{Experimental Setup}

We follow the evaluation protocol of our baseline~\cite{zanella2024delving} by conducting experiments on ShanghaiTech~\cite{shanghaitech}, UCF-Crime~\cite{ucfcrime}, and XD-Violence~\cite{wu2020not}. We also follow the metrics by reporting frame-level AUC and AP for video anomaly detection (VAD), while recording mAUC and mAP for video anomaly recognition (VAR). For the loss weight terms in equation \ref{eq:loss}, we use $\lambda_{\text{sm}}=0.008$ and $\lambda_{\text{sp}}=0.012$ for ShanghaiTech, $\lambda_{\text{sm}}=0.016$ and $\lambda_{\text{sp}}=0.004$ for UCF-Crime, and $\lambda_{\text{sm}}=0.002$ and $\lambda_{\text{sp}}=0.008$ for XD-Violence. We set the number of clusters
K to 32 for UCF-Crime, 20 for XD-Violence, and 2 for ShanghaiTech, the smaller value reflecting ShanghaiTech's substantially shorter average video length compared to the other two datasets.

\subsection{Results}


We compare our method against representative weakly supervised VAD approaches, using AnomalyCLIP as our primary baseline. As shown in Table~\ref{tab:main-results}, our method achieves consistent improvements in binary detection across all three benchmarks: AUC increases from 86.36 to 87.42 on UCF-Crime and from 98.07 to 98.52 on ShanghaiTech, while AP increases from 78.51 to 80.39 on XD-Violence.

For multi-class recognition, our method improves the mAUC on both UCF-Crime (90.66 $\rightarrow$ 91.12) and ShanghaiTech (96.46 $\rightarrow$ 96.74). Notably, the per-class variance $\sigma^2$ decreases substantially on both datasets---from 31.37 to 19.88 on UCF-Crime and from 7.75 to 3.66 on ShanghaiTech---indicating that our method produces more consistent recognition across anomaly classes, rather than excelling partially while underperforming on the rest. Fig.~\ref{fig:qualitative} illustrates representative frame-level anomaly score predictions on UCF-Crime, showing that our method consistently and precisely assigns elevated scores to anomalous segments while suppressing false activations in normal videos.

On XD-Violence, although AP improves, the mAP is marginally lower than that of AnomalyCLIP (49.41 $\rightarrow$ 48.57). The per-class analysis in Table~\ref{tab:xdviolence-class-results} shows that this degradation is driven mainly by Abuse, Shooting, and Fighting, where anomaly recognition depends more on the instantaneous action itself than on broader scene context. This behavior reflects an architectural characteristic of our framework: the proposed modules operate exclusively within Branch~2, refining only the binary anomaly score $\mathbf{s}_{\mathrm{final}}$, while the class-conditional logits in Branch~1 remain unchanged from the baseline. The multi-class score $\mathbf{P}_{\mathrm{class}}$ is obtained by weighting the Branch~1 outputs with $\mathbf{s}_{\mathrm{final}}$; consequently, any miscalibration in $\mathbf{s}_{\mathrm{final}}$ for action-centric classes propagates directly into the final mAP. We also note that lower variance does not necessarily indicate better performance: on XD-Violence, CLIP zero-shot achieves the lowest $\sigma^2$ (498.84) among all compared methods, yet records the lowest mAP (21.32). On UCF-Crime and ShanghaiTech, by contrast, our method simultaneously achieves the highest mAUC and the lowest $\sigma^2$, confirming that the variance reduction reflects genuine consistency rather than uniformly low performance.

\section{Conclusion}
We presented a weakly supervised VAD framework that addresses the inability to group semantically related segments into coherent representations and the lack of scene-context awareness in anomaly scoring. Our proposed HCAT organizes video segments into content-adaptive clusters to jointly represent temporally coherent anomalous patterns and propagate global context back to individual frames. The SDC module complements this by explicitly measuring the compatibility between stable background scene descriptors and local motion, amplifying scores where foreground activity is inconsistent with its scene context. Together, these modules yield consistent improvements in binary detection across three benchmarks, and reduce inter-class performance variance in multi-class recognition on UCF-Crime and ShanghaiTech. 
Strengthening the interaction between temporal modeling and class-conditional recognition represents a direction for future work.

\section*{Acknowledgment}

This work was supported by the ICT R\&D program of MSIT/IITP [RS-2024-00336663, Development of AI technology for tracking and investigating drug criminals through multimedia sources] and by the BK21 FOUR project (AI-driven Convergence Software Education Research Program) funded by the Ministry of Education, School of Computer Science and Engineering, Kyungpook National University, Korea (4120240214871).

\end{document}